\documentclass[10pt,twocolumn,letterpaper]{article}
\usepackage[pagenumbers]{cvpr}
\usepackage{microtype}
\usepackage{needspace}
\usepackage{xspace}
\usepackage{mathtools}
\usepackage{array,multirow,longtable}
\usepackage{colortbl}
\usepackage{tikz}
\usetikzlibrary{arrows.meta,positioning,fit,calc,patterns}
\usepackage{cuted}
\definecolor{cvprblue}{HTML}{357DA3}
\definecolor{ink}{HTML}{44546A}
\definecolor{ours}{HTML}{137C80}
\definecolor{prior}{HTML}{8194A6}
\definecolor{warm}{HTML}{B67B43}
\newcommand{\method}{CoDiCount\xspace}
\newcommand{\R}{\mathbb{R}}
\newcommand{\N}{\mathbb{N}}

\DeclareMathOperator{\softmax}{softmax}
\DeclareMathOperator*{\argmax}{arg\,max}
\DeclareMathOperator*{\argmin}{arg\,min}
\tikzset{box/.style={draw=ink!55,rounded corners=2pt,align=center,font=\sffamily\footnotesize,text=ink,minimum height=9mm,inner sep=4pt},
  newbox/.style={box,draw=ours,fill=ours!7},oldbox/.style={box,fill=prior!10},
  flow/.style={-{Stealth[length=1.6mm]},draw=ink,line width=.65pt}}

\usepackage[pagebackref,breaklinks,colorlinks,allcolors=cvprblue]{hyperref}
\def\confName{CVPR}
\def\confYear{2027}
\title{Counting Beyond Frames:\\Codec Motion Tells Frozen Diffusion VLMs How Many}
\author{Zhuoqin Li\textsuperscript{1,2}\quad Jiayu Ding\textsuperscript{1}\quad Ge Li\textsuperscript{1}\thanks{Corresponding author.}\\[3pt]
\normalsize\textsuperscript{1}Peking University Shenzhen Graduate School, Peking University\\
\normalsize\textsuperscript{2}Harbin Institute of Technology, Shenzhen}
\hypersetup{pdfauthor={Zhuoqin Li, Jiayu Ding, Ge Li}}

\begin{document}
\maketitle
\begin{strip}
\centering
\includegraphics[width=\textwidth]{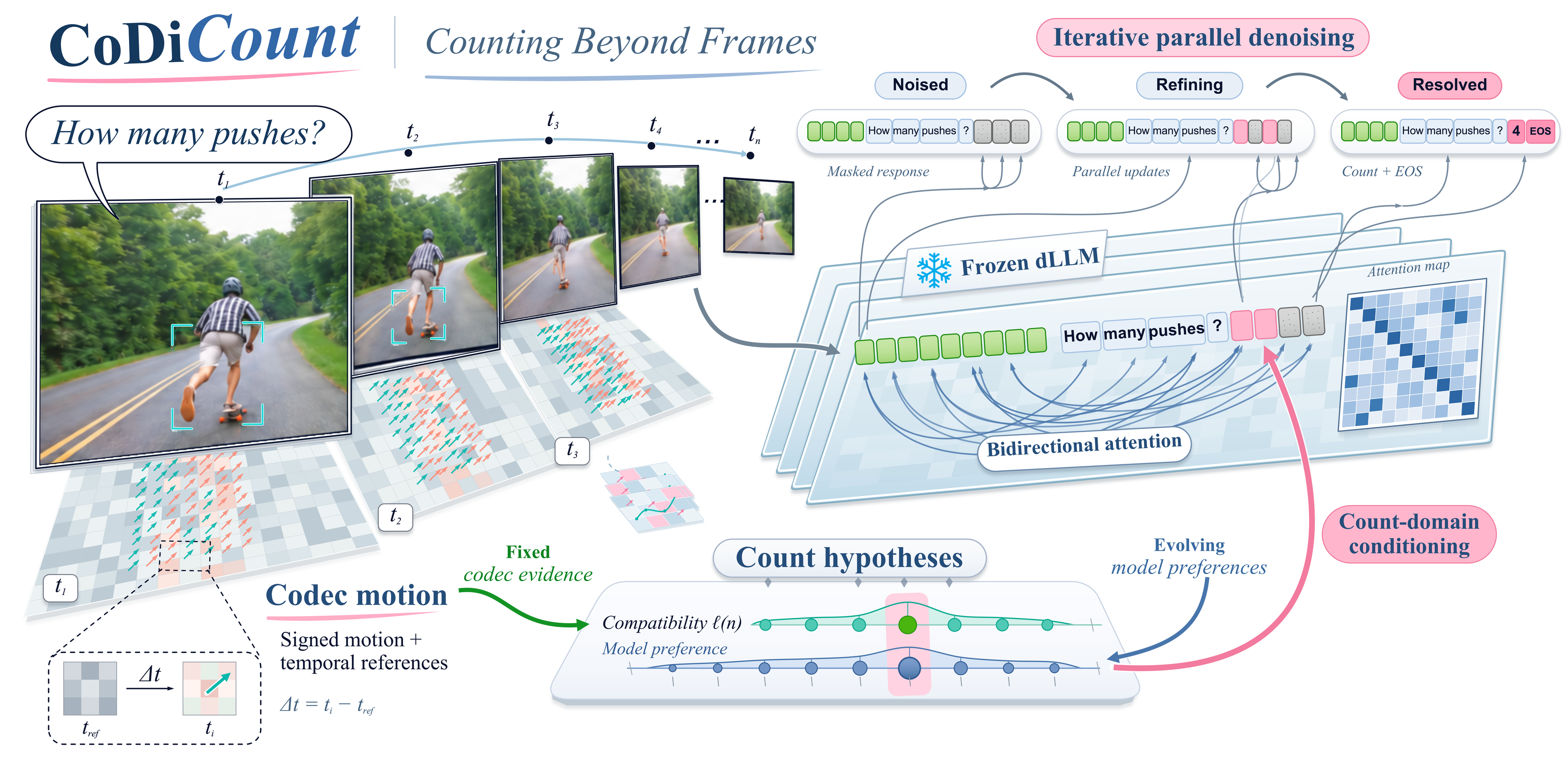}
\captionsetup{hypcap=false}
\captionof{figure}{\textbf{Codec motion tells a frozen diffusion VLM how many.} Video codecs record the movement between the sampled frames, and \method turns this motion into a count potential $\ell(n)$ that tests every candidate count. During denoising, this evidence conditions the evolving answer of a frozen dVLM and steers its unresolved tokens toward the supported count before they are committed.}
\label{fig:teaser}
\end{strip}

\begin{abstract}
Counting how many times something happens is among the simplest questions about a video and among the hardest for video-language models, autoregressive and diffusion alike.
The difficulty lies between the frames: a model samples only a few of them, while each repetition unfolds as movement in between.
Video codecs already record this movement densely as block displacements, and codec-based video models use it to shape their visual input.
We are the first to show that this motion provides direct evidence for the count: a candidate count is a testable motion hypothesis, and unwinding the cycles it implies measures how strongly the observed directions support it.
We present \method, a unified training-free framework that turns this test into a count potential and applies it while a frozen diffusion vision-language model (dVLM) denoises its answer.
At every denoising step, structured count messages map each option letter, or each grammar-valid edit of an open integer, to the count it expresses and reweight the model's own probabilities within that set.
Consistent gains over the official decoders on MotionBench, TOMATO, and SVCBench demonstrate the broad generalization of \method across diffusion models, datasets, and diverse counting tasks.
With \method, a 7B dVLM outperforms GPT-5.1 and Gemini-2.5-Pro on MotionBench, surpasses GPT-4o and much larger open models on TOMATO, and matches GPT-5.4 on SVCBench.
\end{abstract}

\section{Introduction}
\label{sec:intro}

Counting how many times something happens is among the simplest questions about a video and among the hardest for video-language models.
Recent benchmarks test it as part of fine-grained motion understanding in MotionBench~\cite{motionbench}, temporal reasoning in TOMATO~\cite{tomato}, and streaming state maintenance in SVCBench~\cite{svcbench}.
On TOMATO, GPT-4o answers only 36.0\% of the counting questions correctly~\cite{tomato}.
On SVCBench, every model evaluated by its authors remains far below the 96.1 of humans on the primary counting metric, and on periodic activities none exceeds 11.1~\cite{svcbench}.

The difficulty lies between the frames.
A video-language model reasons over a few sampled frames~\cite{ding2026dyn}, whereas each repetition of an action is a cycle of movement that unfolds between them.
When an action repeats quickly, consecutive frames can skip whole repetitions, and no frame records how many were skipped.
Dedicated repetition counters recover this temporal structure through temporal self-similarity in RepNet~\cite{repnet}, multi-scale temporal correlation in TransRAC~\cite{transrac}, or exemplar correspondence in ESCounts~\cite{escounts}, and CountLLM~\cite{countllm} brings repetition counting into a language model through multimodal training.
These counters are dedicated models trained to count repeated actions, whereas the counting questions of video-language benchmarks are answered by general models that must also decide what to count.

Video codecs already record the movement between frames.
To compress a video, the encoder predicts image blocks from reference frames and stores a signed displacement, the motion vector, for each prediction.
Recent video models exploit this structure as input: Video-LaVIT~\cite{videolavit} and CoPE-VideoLM~\cite{cope} learn motion tokens, EMA~\cite{ema} aggregates motion vectors within each group of pictures, and OneVision-Encoder~\cite{onevisionencoder} selects visual patches by codec saliency.
In each case, the codec changes what the model sees, so the motion vectors reach the count only through learned visual features.

We show that codec motion can do more than shape the input: it provides direct evidence for the count.
A candidate count is itself a testable hypothesis about the motion.
For a repeated movement, a proposed number of cycles predicts how the observed directions are organized over time.
Unwinding those cycles and measuring the organization of the residual directions tests this prediction without detecting individual events, and assigns every integer a score that we call the \emph{count potential} (\cref{fig:teaser}).

Diffusion vision-language models (dVLMs) such as LaViDa~\cite{lavida}, Dream-VL~\cite{dreamvl}, and VidLaDA~\cite{vidlada} offer a natural place for this evidence.
They produce an answer by iterative denoising.
At every step, the model scores all unresolved answer positions with bidirectional context and commits a subset of them.
A count therefore stays open while the answer forms, and evidence that enters during denoising can still decide it.
An autoregressive model, by contrast, commits the first digit of a count before the remaining digits exist, so evidence about the complete number cannot reach that decision.

We introduce \method, a unified training-free framework for \emph{count-level conditioning} of a frozen dVLM.
\method evaluates the potential on a motion-progress clock that advances with accumulated movement rather than wall time, so the hypothesized cycles follow the measured movement through pauses and changes of speed.
\emph{Structured count messages} then bring the potential into every denoising step by mapping public option letters, or the grammar-valid edits of the current integer draft, to the counts they express and reweighting the model's own probabilities within that set.
The model's recognition thus decides what is counted, and the codec motion tests how many.
In the full Dream-VL implementation, the messages join codec-selected visual memory, temporal contrast, and soft feedback in one inference loop, so the codec shapes both what the model sees and which count it commits.
All model weights stay frozen, and the count messages connect codec measurements directly to the number being generated without adding any forward pass.

Applied to two dVLMs with different decoders, Dream-VL and LaViDa, \method outperforms the official decoders of both models on all three benchmarks, raising MotionBench accuracy from 23.0\% to 57.5\% on Dream-VL and from 23.2\% to 55.2\% on LaViDa, and LaViDa's SVCBench score by half.
On SVCBench periodic activities, it lifts the GPA of Dream-VL from 0.0 to 66.1.
Our contributions are as follows:
\begin{itemize}
 \item We are the first to use codec motion as direct evidence for the count, testing every candidate integer as a cycle hypothesis through motion progress and phase-marginalized directional organization (\cref{sec:progress,sec:angular}).
 \item Our unified training-free framework, \method, brings this evidence into every denoising step of a frozen dVLM through structured count messages (\cref{sec:messages,sec:interface}).
 \item Without any training, \method generalizes broadly across diffusion models, datasets, and diverse counting tasks, and lets a 7B dVLM surpass GPT-5.1 and Gemini-2.5-Pro on MotionBench and GPT-4o on TOMATO, and match GPT-5.4 on SVCBench (\cref{sec:exp}).
\end{itemize}

\section{Related Work}
\label{sec:related}

\paragraph{Repetition counting.}
Class-agnostic repetition counters learn periodicity from annotated or synthetic repetitions: RepNet~\cite{repnet} through temporal self-similarity, TransRAC~\cite{transrac} through multi-scale temporal correlations, context-aware counting~\cite{contextcount} through coarse-to-fine cycle refinement, and ESCounts~\cite{escounts} through exemplar correspondence.
Other counters add audio~\cite{sightsound}, salient poses~\cite{poserac}, or open-vocabulary targets~\cite{ovr}, and CountLLM~\cite{countllm} trains a multimodal language model with periodic representations.
These methods are dedicated counters trained to count the repetitions of an action.
\method instead strengthens the general video-language models that answer counting questions: the question decides what is counted, and codec evidence for every candidate count, obtained without any training, tests how many.

\paragraph{Counting in video-language benchmarks.}
MotionBench~\cite{motionbench}, TOMATO~\cite{tomato}, and SVCBench~\cite{svcbench} test counting within motion understanding, temporal reasoning, and streaming state maintenance, with units that range from repeated cycles to objects, identities, and complete episodes.
Counting remains far from solved on these tests, both for open models trained by video instruction tuning, such as LLaVA-Video~\cite{llavavideo}, and for proprietary models, and \method applies one training-free potential across all of these settings.

\paragraph{Codec representations.}
Compressed-domain recognition uses motion vectors and residuals as efficient inputs, from CoViAR~\cite{coviar} to the flow-supervised motion cues of DMC-Net~\cite{dmcnet}.
Recent video-language models bring codec structure into their visual input: Video-LaVIT~\cite{videolavit} and CoPE-VideoLM~\cite{cope} learn motion tokens, EMA~\cite{ema} aggregates codec features within groups of pictures, and OneVision-Encoder~\cite{onevisionencoder} and AdaCodec~\cite{adacodec} allocate visual tokens from coding structure.
In all of these methods, the codec decides what the model sees.
\method is the first to turn codec motion into direct evidence for the count: its integer-indexed potential scores every numerical hypothesis and acts on the answer of a frozen model.

\paragraph{Diffusion vision-language models.}
Discrete diffusion generates by iterative denoising~\cite{d3pm,mdlm,maskgit}, and LLaDA~\cite{llada} and Dream~\cite{dream} scale it to instruction-following language models.
LaViDa~\cite{lavida}, Dream-VL~\cite{dreamvl}, and VidLaDA~\cite{vidlada} extend it to images and video, keeping every unresolved answer position open across denoising steps.
Training-free decoding methods such as contrastive decoding~\cite{vcd,videocd} and geometry-aware soft tokens~\cite{softtokens} improve how these models use their visual input or commit tokens, and the Dream-VL instantiation of \method builds on both.
\method adds evidence about the count itself, conditioning the evolving token alternatives on the counts they express at every denoising step.

\Needspace{8\baselineskip}
\section{Codec Motion as Count Evidence}
\label{sec:method}

\begin{figure*}[t]
\centering
\includegraphics[width=\textwidth]{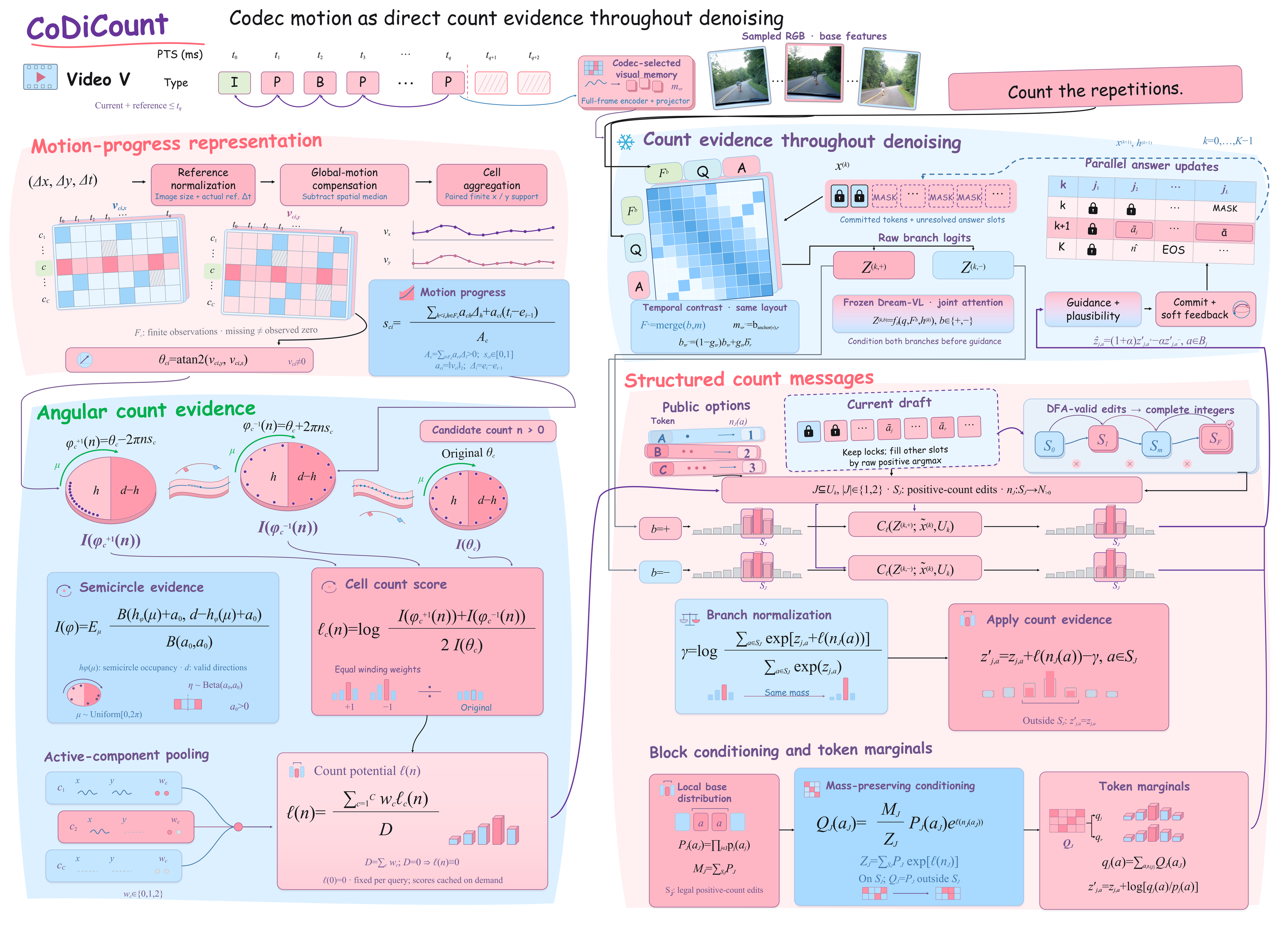}
\caption{\textbf{\method turns codec motion into count evidence inside the denoising of a frozen dVLM.} \textbf{Left:} reference-normalized, motion-compensated cell velocities define directions and motion progress. Unwinding each candidate count in both directions and measuring semicircle organization yields a shared potential $\ell(n)$. \textbf{Right:} public options or DFA-valid edits map the evolving answer draft to complete integers. Normalized count messages condition both visual branches before guidance; coupled numeric and termination edits produce token marginals. Commitment and soft feedback carry the conditioned predictions into the next step of the frozen Dream-VL decoder.}
\label{fig:overview}
\end{figure*}

Video counting determines a compact numerical answer from dense temporal evidence.
Diffusion vision-language models (dVLMs) offer a natural interface for this evidence, because their unresolved answer positions can be conditioned jointly and refined across denoising steps.
\method, a unified training-free framework for \emph{count-level conditioning}, connects the two by keeping codec evidence in \emph{count space} and conditioning the evolving answer on it.
A \emph{count potential} $\ell(n)$ tests each candidate integer against the observed motion (\cref{sec:progress,sec:angular}), and \emph{structured count messages} translate this evidence into coordinated updates of the answer tokens (\cref{sec:messages}).
The inference loop then carries the conditioned predictions through denoising with codec-selected visual memory, temporal guidance, and soft feedback (\cref{sec:interface,fig:overview}).

Given a video and question $q$, we restrict observations to the question's temporal domain and predict either an option letter or a nonnegative integer.
Let $\mathcal{A}$ be a vocabulary of size $V$, $L$ the number of answer slots, and $K$ the denoising budget.
At step $k\in\{0,\ldots,K-1\}$, $\mathbf{x}^{(k)}\in(\mathcal{A}\cup\{\texttt{MASK}\})^L$ records the committed tokens, and $\mathcal{U}_k=\{j:x_j^{(k)}=\texttt{MASK}\}$ contains the unresolved positions.
The potential is fixed by the video and query, while its mapping to token choices follows the evolving answer.
All model weights remain frozen, and one motion statistic and one message rule serve every question category, with the question's temporal domain and answer format supplying the query-specific inputs.

\subsection{Motion-progress representation}
\label{sec:progress}

\paragraph{Signed motion with explicit support.}
Codec motion vectors describe block predictions between a current frame and its references.
We convert their signed displacements to velocities using the actual reference time differences and normalize by image dimensions.
After averaging duplicate measurements, we subtract the component-wise spatial median for each current/reference pair to compensate for global translation and aggregate the remaining motion into a grid of $C$ spatial cells.
This gives a cell velocity $\mathbf{v}_{ci}\in\R^2$ for cell $c\in\{1,\ldots,C\}$ at native time $t_i$, with an explicit mask separating unavailable measurements from observed zero motion.
For a prefix query, both the current frame and every used reference must precede its cutoff.
Appendix~A specifies the spatial reduction and reference handling.

\paragraph{Accumulated movement as a coordinate.}
Elapsed time gives unequal phase increments to repetitions performed at different speeds.
We instead parameterize each cell by its accumulated observed movement.
For $t_1<\cdots<t_m$, define bin edges $e_0=t_1$, $e_m=t_m$, and $e_i=(t_i+t_{i+1})/2$ for $1\leq i<m$.
Let $\Delta_i=e_i-e_{i-1}$, $a_{ci}=\|\mathbf{v}_{ci}\|_2$, and $\mathcal{F}_c$ be the finite observations of cell $c$.
With $A_c=\sum_{i\in\mathcal{F}_c}a_{ci}\Delta_i>0$, its motion progress is
\begin{equation}
 s_{ci}=
 \frac{\displaystyle\sum_{\substack{h<i\\h\in\mathcal{F}_c}}a_{ch}\Delta_h
       +a_{ci}(t_i-e_{i-1})}{A_c}\in[0,1].
 \label{eq:progress}
\end{equation}
The horizontal and vertical components share this coordinate.
Observed pauses add no progress, and equal progress increments represent equal amounts of measured movement.
Missing bins add no observed mass, so the clock advances only with measured movement.
This coordinate aligns count hypotheses by movement rather than duration, without locating individual cycle boundaries.

\subsection{Angular count evidence}
\label{sec:angular}

\paragraph{A count as a winding hypothesis.}
The central idea of \method is that a candidate count is a testable hypothesis about the motion.
For finite nonzero vectors, define the direction $\theta_{ci}=\operatorname{atan2}(v_{ci,y},v_{ci,x})$.
A positive count $n$ induces two residual phase sequences,
\begin{equation}
 \boldsymbol{\phi}_c^{\,\sigma}(n)
 =\boldsymbol{\theta}_c-\sigma\,2\pi n\mathbf{s}_c,
 \qquad \sigma\in\{-1,+1\}.
 \label{eq:phase}
\end{equation}
Each hypothesis asks whether unwinding $n$ cycles makes the directions more organized.
The test covers circular as well as reciprocating motion: under the correct winding, the outgoing and returning directions of a back-and-forth cycle map into a common semicircle.
It measures directional support for each count, while the question and the visual model specify the unit being counted.
Zero vectors contribute zero speed to \cref{eq:progress} and no angle to \cref{eq:phase}.

\paragraph{Marginalizing phase and concentration.}
The starting direction is arbitrary, and a fixed angular concentration would impose a motion-specific scale.
We therefore model directions with a two-semicircle density whose origin $\mu$ and half-circle mass $\eta$ are both marginalized.
For $d$ angles $\boldsymbol{\phi}$, let $h_{\boldsymbol{\phi}}(\mu)$ count those in $(\mu,\mu+\pi]$ modulo $2\pi$.
Integrating $\eta$ under a symmetric prior $\operatorname{Beta}(a_0,a_0)$ and $\mu\sim\operatorname{Uniform}[0,2\pi)$ gives, up to the common density factor $\pi^{-d}$,
\begin{equation}
 I(\boldsymbol{\phi})=\frac{1}{2\pi}\int_0^{2\pi}
 \frac{\mathrm{B}\!\left(h_{\boldsymbol{\phi}}(\mu)+a_0,
 d-h_{\boldsymbol{\phi}}(\mu)+a_0\right)}
 {\mathrm{B}(a_0,a_0)}\,\mathrm{d}\mu .
 \label{eq:integral}
\end{equation}
Here $\mathrm{B}$ is the beta function, and $a_0>0$ sets the prior on the half-circle mass.
The statistic rewards angular organization without a preferred origin, and since its integrand changes only at the observed angles and their $\pi$-shifted values, it is evaluated exactly as a finite sum.

We compare the equal mixture of both winding directions with the original, unwound directions:
\begin{equation}
 \ell_c(n)=\log
 \frac{I(\boldsymbol{\phi}_c^{+1}(n))
       +I(\boldsymbol{\phi}_c^{-1}(n))}
      {2I(\boldsymbol{\theta}_c)} .
 \label{eq:cell}
\end{equation}
The original directions set the zero point of the score, so positive values mark counts that organize the motion beyond its raw directions.
A velocity component is active when it retains non-negligible residual energy after fitting a constant and a linear function of source time.
Let $w_c\in\{0,1,2\}$ count the active components of cell $c$, out of its horizontal and vertical ones, and let $D=\sum_c w_c$.
The shared count potential is the weighted mean
\begin{equation}
 \ell(n)=\frac{1}{D}\sum_{c=1}^{C}w_c\ell_c(n),
 \qquad D>0.
 \label{eq:potential}
\end{equation}
Averaging pools the active components on a common scale, with the same active set for every candidate count.
An interval without active components gives the neutral potential $\ell(n)\equiv0$, and we set $\ell(0)=0$ as the reference for positive-count alternatives.
The potential is the numerical interface between the motion representation and the evolving answer: its motion coordinates and active set are prepared once per query, and candidate scores are computed on demand and cached as the answer evolves.

\subsection{Structured count messages}
\label{sec:messages}

The dVLM predicts vocabulary distributions, whereas $\ell$ scores integers.
We bridge this gap with one local conditioning operator, applied to each condition branch at each step.

\paragraph{Conditioning a block of answer positions.}
Suppress the step and branch indices, and let $p_j(a)=\softmax(\mathbf{z}_j)_a$ be the raw token probabilities.
For a block $J\subseteq\mathcal{U}_k$ of one or two positions, form the local base distribution
$P_J(\mathbf{a}_J)=\prod_{j\in J}p_j(a_j)$.
The current draft defines a support $\mathcal{S}_J$ of edits expressing positive integers and a mapping $n_J:\mathcal{S}_J\to\N_{>0}$.
Let $M_J=\sum_{\mathcal{S}_J}P_J$ and
$Z_J=\sum_{\mathcal{S}_J}P_J\exp\ell(n_J)$.
Our count-conditioned distribution is
\begin{equation}
 Q_J(\mathbf{a}_J)=
 \begin{cases}
 \displaystyle\frac{M_J}{Z_J}P_J(\mathbf{a}_J)
       e^{\ell(n_J(\mathbf{a}_J))},
       &\mathbf{a}_J\in\mathcal{S}_J,\\[2pt]
 P_J(\mathbf{a}_J),&\mathbf{a}_J\notin\mathcal{S}_J .
 \end{cases}
 \label{eq:conditioning}
\end{equation}
This local update combines the model's semantic preference with codec count evidence.
It redistributes the existing mass $M_J$ among competing counts, so the model keeps deciding how likely a count answer is at this position while the codec evidence decides which count.
Empty supports and constant potentials leave the distribution unchanged, and Appendix~C derives these properties together with an equivalent constrained-KL form.

To return to the denoiser's token interface, take the marginals
$q_j(a)=\sum_{\mathbf{a}_{J\setminus\{j\}}}Q_J(\mathbf{a}_J)$ and send
\begin{equation}
 z'_{j,a}=z_{j,a}+\log\frac{q_j(a)}{p_j(a)},
 \qquad j\in J.
 \label{eq:message_projection}
\end{equation}
For a single position, this reduces to a normalized additive message:
\begin{align}
 \gamma&=\log\frac{\sum_{a\in\mathcal{S}_J}
                         e^{z_{j,a}+\ell(n_J(a))}}
                        {\sum_{a\in\mathcal{S}_J}e^{z_{j,a}}},
 \label{eq:normalizer}\\
 z'_{j,a}&=
 \begin{cases}
 z_{j,a}+\ell(n_J(a))-\gamma,&a\in\mathcal{S}_J,\\
 z_{j,a},&a\notin\mathcal{S}_J .
 \end{cases}
 \label{eq:unary}
\end{align}
The unary update thus preserves the full softmax normalizer, the total positive-count mass, and every outside probability.
For two positions, \cref{eq:message_projection} turns the joint count-conditioned distribution into the two token distributions that the denoiser consumes.

\paragraph{Domains induced by the answer representation.}
For multiple-choice counting, $J$ contains the first answer position, and the public options map letter tokens to counts.
Only letters expressing positive counts belong to $\mathcal{S}_J$.

For open integers, we construct a complete draft $\widetilde{\mathbf{x}}^{(k)}$ by retaining committed tokens and filling unresolved positions with the positive branch's raw argmax.
A deterministic finite automaton for integer-answer syntax~\cite{outlines} maps legal token edits of this full draft to their integer values.
The automaton supplies the count of each edit for \cref{eq:conditioning}.
The two branches share this map, each with its own $P_J$, and token aliases remain separate choices with their own probabilities.

Since the value of an integer also depends on where the answer ends, the block for the first unresolved numeric position also includes the unresolved end position, and all other unresolved positions use single-position blocks.
Committed positions are never reopened, and Appendix~C details the edit domains and the tokenizer mapping.

\subsection{Count evidence throughout denoising}
\label{sec:interface}

The full Dream-VL instantiation integrates the count messages with codec-selected observations, temporal contrast, and feedback to unresolved positions, all driven by the same video evidence and answer state.

\paragraph{Visual memory and temporal contrast.}
The $T$ sampled RGB frames provide base projected features $\mathbf{b}_{\tau r}$, indexed by tubelet $\tau$ and spatial patch $r$.
At additional native times $\nu$, codec motion magnitudes select up to $N_{\mathrm{mem}}$ supplementary patches.
Each selected feature $\mathbf{m}_{\nu r}$ comes from full-frame encoding by the same frozen vision encoder and projector, adding image context at native times between the sampled frames.

The positive condition $\mathbf{F}^{+}$ merges base and supplementary features in temporal order.
For the negative condition, define
\begin{equation}
 \begin{aligned}
 \mathbf{b}^{-}_{\tau r}
   &=(1-g_{\tau r})\mathbf{b}_{\tau r}
       +g_{\tau r}\overline{\mathbf{b}}_r,\\
 \mathbf{m}^{-}_{\nu r}
   &=\mathbf{b}_{\operatorname{anchor}(\nu),r},
 \end{aligned}
 \label{eq:conditions}
\end{equation}
where $\overline{\mathbf{b}}_r$ is the temporal mean at patch $r$,
$g_{\tau r}\in[0,1]$ is the fixed codec-motion gate, and
$\operatorname{anchor}(\nu)$ is the nearest base tubelet.
Merging these features in the same positions yields $\mathbf{F}^{-}$.
The negative condition thus attenuates temporal variation while keeping the token layout of the positive one, so the two predictions are directly comparable at the same answer state.

\paragraph{Count conditioning before guidance.}
Let $\mathbf{h}^{(k)}$ denote the answer embeddings, shared by both branches.
The frozen denoiser $f_{\vartheta}$ and the message operator $\mathcal{C}_{\ell}$ give
\begin{equation}
 \begin{aligned}
 \mathbf{Z}^{(k,b)}
   &=f_{\vartheta}(q,\mathbf{F}^{b},\mathbf{h}^{(k)}),\\
 \mathbf{Z}^{\prime(k,b)}
   &=\mathcal{C}_{\ell}
       (\mathbf{Z}^{(k,b)};\widetilde{\mathbf{x}}^{(k)},\mathcal{U}_k),
 \quad b\in\{+,-\}.
 \end{aligned}
 \label{eq:branch_update}
\end{equation}
Following visual contrastive decoding~\cite{videocd}, we use
\begin{equation}
 \widehat z^{(k)}_{j,a}=
 \begin{cases}
 (1+\alpha)z^{\prime(k,+)}_{j,a}
       -\alpha z^{\prime(k,-)}_{j,a},&a\in\mathcal{B}^{(k)}_j,\\
 -\infty,&a\notin\mathcal{B}^{(k)}_j ,
 \end{cases}
 \label{eq:guidance}
\end{equation}
where $\alpha>0$ is the contrast strength and $\mathcal{B}^{(k)}_j=\{a:z^{\prime(k,+)}_{j,a}\geq
\max_u z^{\prime(k,+)}_{j,u}+\log\beta\}$ keeps tokens within a plausibility ratio $\beta\in(0,1]$ of the most likely one.
When the two conditions coincide, the host uses the positive logits directly.
Conditioning both branches with the same unary potential leaves exactly one copy of $\ell(n)$ after contrast, because $(1+\alpha)\ell-\alpha\ell=\ell$, while each branch keeps its own count mass before guidance.

\paragraph{Carrying count evidence between steps.}
The native commitment schedule selects tokens from the guided logits.
For still-unresolved slots, we also retain distributional information through geometry-aware soft tokens~\cite{softtokens}.
Let $\mathcal{T}^{(k)}_j$ contain the $m_{\mathrm{s}}$ largest guided logits, and let $\pi_j^{(k)}$ be their softmax normalized within that set.
With token embedding $\mathbf{E}_a$, define
$\boldsymbol{\mu}^{(k)}_j=\sum_{a\in\mathcal{T}^{(k)}_j}\pi_j^{(k)}(a)\mathbf{E}_a$.
The next input is
\begin{equation}
 \mathbf{h}^{(k+1)}_j=
 \begin{cases}
 \mathbf{E}_{x_j^{(k+1)}},&j\notin\mathcal{U}_{k+1},\\[2pt]
 \rho\,\operatorname{slerp}\!\left(
 \widehat{\mathbf{E}}_{\texttt{MASK}},
 \widehat{\boldsymbol{\mu}}^{(k)}_j;
 \lambda\right),&j\in\mathcal{U}_{k+1},
 \end{cases}
 \label{eq:feedback}
\end{equation}
where $\widehat{\cdot}$ denotes unit normalization, $\rho=\|\mathbf{E}_{\texttt{MASK}}\|_2$, $\operatorname{slerp}$ interpolates along the unit sphere, and $\lambda\in[0,1]$ sets how far the proposal moves away from MASK.
Initialization uses MASK embeddings at every position.
The count-conditioned distribution therefore shapes both the current commitment and the next denoiser input, and each new draft updates the legal count edits against the same codec potential.

\method uses the same motion potential and structured count messages in Dream-VL and LaViDa (\cref{sec:setup}).

\section{Experiments}
\label{sec:exp}

We evaluate count-level conditioning across diffusion models with different decoders, across benchmarks with multiple-choice and open-integer answers, and across counting categories, and then ablate the parts of the count evidence.
A single training-free configuration of the count potential and messages serves all benchmarks, categories, and both models.

\subsection{Evaluation protocol}
\label{sec:setup}

\paragraph{Benchmarks.}
We evaluate the public Repetition Count questions of MotionBench~\cite{motionbench}, the Action Count questions of TOMATO~\cite{tomato}, and the complete SVCBench~\cite{svcbench} with all eight counting categories.
MotionBench and TOMATO are multiple-choice with four to six options and span gaming, industrial, medical, sports, human, object, and simulated videos.
SVCBench asks for an open integer after every prefix of a stream and covers event counting (actions, transitions, episodes, and periodic activities) and object counting.

\paragraph{Models and baselines.}
We apply \method to two diffusion vision-language models with different backbones, vision encoders, and decoders, and compare each with its official decoder.
For Dream-VL-7B~\cite{dreamvl}, the baseline is the unmodified official generator.
For LaViDa~\cite{lavida}, both arms run the model's native hard-commit decoder with 32 frames, 32 answer slots, and 16 denoising steps, and \method adds only the count potential and messages to its answer logits.
On Dream-VL, \method runs the complete inference loop of \cref{sec:interface} with 32 language-model calls per answer, and on LaViDa it uses the native 17 calls of its host.

\paragraph{Implementation.}
The motion grid has $C=4\times4$ cells, and the angular statistic uses the Jeffreys prior $a_0=1/2$.
On Dream-VL, \method uses $L=32$ answer slots, $K=16$ denoising steps, and $T=16$ sampled frames with a visual memory of $N_{\mathrm{mem}}=512$ patches, together with the fixed host settings $\alpha=1$ and $\beta=0.2$ of \cite{videocd} and $m_{\mathrm{s}}=3$ and $\lambda=0.1$ of \cite{softtokens}.
Tokens are committed by each host's confidence-based schedule at zero temperature, and for SVCBench every sampled frame and codec reference precedes the query time.

\paragraph{Metrics.}
MotionBench and TOMATO use accuracy over all questions.
SVCBench uses its official record-level \textit{Gaussian} Precision Accuracy (GPA), which averages over the queries of each record and then over records.
For $R\in\N$ records, $m_r\in\N$ query points in record $r$, prediction $\hat n_{rj}\in\mathbb{Z}$, and reference count $n_{rj}\in\N_0$, GPA is
\begin{equation}
 \operatorname{GPA}=\frac1R\sum_{r=1}^R\frac1{m_r}\sum_{j=1}^{m_r}
 \exp\!\left[-\frac{(\hat n_{rj}-n_{rj})^2}
 {2\{0.05\max(n_{rj},1)\}^2}\right].
 \label{eq:gpa}
\end{equation}
Following the official protocol, we extract one explicit integer from each answer and score an answer without one as incorrect.

\subsection{Counting across models and benchmarks}
\label{sec:mainresults}

\begin{table}[t]
\centering
\caption{\textbf{MotionBench.} With \method, the 7B Dream-VL surpasses GPT-5.1 and Gemini-2.5-Pro. Accuracy (\%). Autoregressive results are from the official MotionBench leaderboard~\cite{motionbench} on the same questions. Shaded rows are ours, with the gain over the official decoder.}
\label{tab:motionbenchar}
\footnotesize
\setlength{\tabcolsep}{4pt}
\begin{tabular}{lcc}
\toprule
Model & Size & Accuracy \\
\midrule
\multicolumn{3}{l}{\textit{Proprietary autoregressive models}} \\
Gemini-2.5-Pro & -- & 57.0 \\
GPT-5.1 & -- & 56.7 \\
Claude-Opus-4.5 & -- & 36.9 \\
Claude-Sonnet-4.5 & -- & 29.4 \\
GPT-5-Chat & -- & 26.7 \\
Seed1.5-VL & -- & 26.7 \\
\midrule
\multicolumn{3}{l}{\textit{Open-source autoregressive models}} \\
GLM-4.5V-106B & 106B & 40.4 \\
InternVL2.5-78B & 78B & 33.7 \\
Qwen2.5-VL-72B & 72B & 28.7 \\
\midrule
\multicolumn{3}{l}{\textit{Diffusion models}} \\
Dream-VL & 7B & 23.0 \\
\rowcolor{ours!8}\quad + \method & 7B & \textbf{57.5\,{\tiny\textcolor{ours}{+34.5}}} \\
LaViDa & 8B & 23.2 \\
\rowcolor{ours!8}\quad + \method & 8B & 55.2\,{\tiny\textcolor{ours}{+32.0}} \\
\bottomrule
\end{tabular}
\end{table}

\begin{table}[t]
\centering
\caption{\textbf{TOMATO.} With \method, the 7B Dream-VL surpasses every listed model. Accuracy (\%). Autoregressive results are reported by the TOMATO authors~\cite{tomato} on the same questions. Shaded rows are ours, with the gain over the official decoder.}
\label{tab:tomato}
\footnotesize
\setlength{\tabcolsep}{4pt}
\begin{tabular}{lcc}
\toprule
Model & Size & Accuracy \\
\midrule
\multicolumn{3}{l}{\textit{Proprietary autoregressive models}} \\
GPT-4o & -- & 36.0 \\
Gemini-1.5-Pro & -- & 36.3 \\
Claude-3.5-Sonnet & -- & 31.2 \\
\midrule
\multicolumn{3}{l}{\textit{Open-source autoregressive models}} \\
LLaVA-Video-72B & 72B & 34.6 \\
VideoLLaMA2-72B & 72B & 28.8 \\
InternVL2-40B & 40B & 29.1 \\
LLaVA-NeXT-Video-32B & 32B & 24.3 \\
InternVL2-26B & 26B & 25.7 \\
Video-CCAM-v1.1-14B & 14B & 35.6 \\
VILA-13B & 13B & 27.7 \\
\midrule
\multicolumn{3}{l}{\textit{Diffusion models}} \\
Dream-VL & 7B & 31.5 \\
\rowcolor{ours!8}\quad + \method & 7B & \textbf{43.2\,{\tiny\textcolor{ours}{+11.6}}} \\
LaViDa & 8B & 17.8 \\
\rowcolor{ours!8}\quad + \method & 8B & 30.5\,{\tiny\textcolor{ours}{+12.7}} \\
\bottomrule
\end{tabular}
\end{table}

\begin{table}[t]
\centering
\caption{\textbf{SVCBench.} With \method, the 7B Dream-VL matches GPT-5.4 overall and leads every listed model on periodic activities. Official GPA$\times$100 on the complete benchmark (Overall) and on its periodic activities (Periodic). Autoregressive results are reported by the SVCBench authors~\cite{svcbench} on the same queries, with 64 frames or one frame per second. Size denotes total parameters. Shaded rows are ours, with the gain over the official decoder.}
\label{tab:svcbench}
\footnotesize
\setlength{\tabcolsep}{4pt}
\begin{tabular}{lccc}
\toprule
 & & \multicolumn{2}{c}{GPA} \\
\cmidrule(lr){3-4}
Model & Size & Overall & Periodic \\
\midrule
\multicolumn{4}{l}{\textit{Frontier autoregressive models}} \\
GPT-5.4 & -- & 29.1 & 3.2 \\
Kimi-K2.5 & 1T & 26.4 & 0.4 \\
\midrule
\multicolumn{4}{l}{\textit{Open-source autoregressive models}} \\
Qwen3-VL-30B & 30B & 27.0 & 2.5 \\
InternVL-3.5-8B & 8B & 24.2 & 1.0 \\
\midrule
\multicolumn{4}{l}{\textit{Streaming autoregressive models}} \\
Flash-VStream-7B & 7B & 23.4 & 0.0 \\
LiveStar & -- & 19.9 & 11.1 \\
\midrule
\multicolumn{4}{l}{\textit{Diffusion models}} \\
Dream-VL & 7B & 26.4 & 0.0 \\
\rowcolor{ours!8}\quad + \method & 7B & 29.1\,{\tiny\textcolor{ours}{+2.7}} & \textbf{66.1\,{\tiny\textcolor{ours}{+66.1}}} \\
LaViDa & 8B & 15.1 & 0.0 \\
\rowcolor{ours!8}\quad + \method & 8B & 22.8\,{\tiny\textcolor{ours}{+7.7}} & 26.6\,{\tiny\textcolor{ours}{+26.6}} \\
\bottomrule
\end{tabular}
\end{table}

\method consistently outperforms the official decoders of both models on all three benchmarks.
On Dream-VL, accuracy rises from 23.0\% to 57.5\% on MotionBench and from 31.5\% to 43.2\% on TOMATO, and SVCBench GPA rises from 26.4 to 29.1.
On LaViDa, accuracy rises from 23.2\% to 55.2\% on MotionBench and from 17.8\% to 30.5\% on TOMATO, and SVCBench GPA rises from 15.1 to 22.8.
The same count potential and messages thus serve a guided decoder with soft feedback and a native hard-commit decoder, without any change.

Without any training, \method lets 7B and 8B diffusion models count at the level of much larger and newer autoregressive models.
On MotionBench, Dream-VL with \method reaches 57.5\%, above Gemini-2.5-Pro (57.0\%), GPT-5.1 (56.7\%), and Claude-Opus-4.5 (36.9\%) on the official leaderboard, and LaViDa with \method reaches 55.2\%, above every listed open model up to 106B (\cref{tab:motionbenchar}).
On TOMATO, Dream-VL with \method reaches 43.2\%, above every model in \cref{tab:tomato}, including Gemini-1.5-Pro (36.3\%), GPT-4o (36.0\%), and LLaVA-Video-72B~\cite{llavavideo} (34.6\%).
On SVCBench, its GPA of 29.1 matches GPT-5.4 (29.1) and exceeds Kimi-K2.5 (26.4), Qwen3-VL-30B (27.0), InternVL-3.5-8B (24.2), and the streaming models (\cref{tab:svcbench}).
The lead is largest on periodic activities, where the count is a number of repeated movements: Dream-VL with \method reaches a GPA of 66.1, against 3.2 for GPT-5.4, and exceeds the strongest evaluated model, LiveStar (11.1), by 55.0.
These gains come from evidence that the sampled frames do not contain: the motion between them, tested against every candidate count.

\begin{table}[t]
\centering
\caption{\textbf{MotionBench by category.} \method improves every category on both models. Accuracy (\%) overall and by official category. Each model is compared with its official decoder, and small numbers give the gain of \method.}
\label{tab:motionbench}
\footnotesize
\setlength{\tabcolsep}{3.5pt}
\begin{tabular}{l c >{\columncolor{ours!8}}c c >{\columncolor{ours!8}}c}
\toprule
 & \multicolumn{2}{c}{Dream-VL (7B)} & \multicolumn{2}{c}{LaViDa (8B)} \\
\cmidrule(lr){2-3}\cmidrule(lr){4-5}
Category & Official & + \method & Native & + \method \\
\midrule
Gaming & 21.7 & 90.8\,{\tiny\textcolor{ours}{+69.1}} & 27.0 & 93.4\,{\tiny\textcolor{ours}{+66.4}} \\
Industrial & 24.2 & 30.1\,{\tiny\textcolor{ours}{+5.9}} & 21.6 & 27.5\,{\tiny\textcolor{ours}{+5.9}} \\
Medical & 22.6 & 49.1\,{\tiny\textcolor{ours}{+26.4}} & 26.4 & 43.4\,{\tiny\textcolor{ours}{+17.0}} \\
Sports & 16.0 & 52.0\,{\tiny\textcolor{ours}{+36.0}} & 20.0 & 48.0\,{\tiny\textcolor{ours}{+28.0}} \\
\midrule
\textbf{All} & 23.0 & 57.5\,{\tiny\textcolor{ours}{+34.5}} & 23.2 & 55.2\,{\tiny\textcolor{ours}{+32.0}} \\
\bottomrule
\end{tabular}
\end{table}

\subsection{Generalization across counting categories}
\label{sec:transfer}

The gains generalize across counting categories.
On MotionBench, every category improves on both models (\cref{tab:motionbench}), and the largest gains occur where the counted unit is a repeated movement: Gaming rises from 21.7\% to 90.8\% on Dream-VL and from 27.0\% to 93.4\% on LaViDa.
On TOMATO, human actions rise from 14.0\% to 70.2\% and from 3.5\% to 45.6\%, and on SVCBench, periodic activities, which neither official decoder counts correctly, reach a GPA of 66.1 on Dream-VL and 26.6 on LaViDa.
Together with the transfer from Dream-VL to LaViDa, these results demonstrate the broad generalization of count-level conditioning across models, answer formats, and counting categories.

\begin{table}[t]
\centering
\caption{\textbf{Ablations of \method on Dream-VL.} All variants share the frozen model, inputs, inference loop, and 32 language-model calls per answer. The first row removes the count potential and structured count messages, and the other rows each change one part of \method on the full SVCBench. MotionBench (MB) and TOMATO (TO) report accuracy (\%). SVCBench reports GPA, Monotonicity Consistency (MoC), and Update Direction Accuracy (UDA), all scaled by 100.}
\label{tab:ablation}
\footnotesize
\setlength{\tabcolsep}{3pt}
\begin{tabular}{lccccc}
\toprule
Variant & MB & TO & GPA & MoC & UDA \\
\midrule
w/o count messages & 26.0 & 28.1 & 24.1 & 67.7 & 33.6 \\
Shuffled codec motion & -- & -- & 24.2 & 58.8 & 38.3 \\
Last-step injection & -- & -- & 25.8 & 64.0 & 35.0 \\
Single-token edits & -- & -- & 26.2 & 66.0 & 37.2 \\
\midrule
\method & \textbf{57.5} & \textbf{43.2} & \textbf{29.1} & \textbf{68.9} & \textbf{39.2} \\
\bottomrule
\end{tabular}
\end{table}

\subsection{Ablations}
\label{sec:ablation}

We ablate \method on Dream-VL, where all variants share the frozen model, the inputs, the inference loop, and 32 language-model calls per answer (\cref{tab:ablation}).

\paragraph{Count evidence.}
Removing the count potential and structured count messages from the complete loop lowers accuracy from 57.5\% to 26.0\% on MotionBench and from 43.2\% to 28.1\% on TOMATO, and SVCBench GPA from 29.1 to 24.1.
The count evidence thus drives the gains of \method within its inference loop.

\paragraph{Temporal order of the codec motion.}
The count potential reads cycles from the order of the motion.
Permuting the codec vectors over time, while keeping their values, timestamps, and support, lowers GPA to 24.2, the level without count messages, and MoC from 68.9 to 58.8.
This confirms the central idea of \method: the count evidence lies in the temporal organization of the codec motion, not in its amount or the distribution of its directions.

\paragraph{Evidence during iterative denoising.}
Applying the same messages only once, at the last denoising step, when most answer positions are already committed, lowers GPA to 25.8.
Count evidence must therefore enter while the answer is still open, as it does at every step of the iterative denoising of a dVLM.

\paragraph{Edit domain of the messages.}
Restricting the messages to single-token edits lowers GPA to 26.2 overall and from 66.1 to 16.7 on periodic activities, whose larger counts often need several answer tokens to change at once.
The structured messages are therefore what make large open-integer counts reachable.

\subsection{Efficiency and extensions}
\label{sec:boundary}

The count potential and messages are lightweight: they add no language-model or vision-encoder forward passes to their host.
Codec fields are exported once per video and reused by every query on it, and the export compresses raw codec data by roughly 500 times into compact fields.
The count potential is computed once per query, and the messages operate on the logits of the count-bearing positions only.

The count potential measures cycles of motion direction, while the question and the dVLM's visual interpretation decide what is counted.
The same count-level interface can host potentials built from other codec signals, such as residuals or block types, to bring identity and grouping evidence into the answer as well.

\Needspace{7\baselineskip}
\section{Conclusion}
\label{sec:conclusion}

We have shown, for the first time, that the motion recorded by video codecs between sampled frames provides direct evidence for video counting at the level of the generated answer.
\method tests each candidate count against codec motion on a motion-progress clock and brings the resulting potential into denoising through messages on the tokens that express each count.
Without any training, this count-level conditioning generalizes broadly across diffusion models, datasets, and diverse counting tasks.
It lets a 7B dVLM count better than GPT-5.1 and Gemini-2.5-Pro on MotionBench and GPT-4o on TOMATO, and on par with GPT-5.4 on SVCBench, where it leads every evaluated model on periodic activities.
Its gains come from the temporal order of the codec motion and from evidence that takes part in the iterations that form the answer.
These results establish the codec as a new source of evidence for video understanding, and the same count-level interface can host potentials from other codec signals, such as residuals and block types, to capture identity and event structure.

{\small\bibliographystyle{ieeenat_fullname}\bibliography{codicount-references}}
\twocolumn[\begin{center}\Large\bfseries Appendix\end{center}]
\appendix
\numberwithin{equation}{section}
\section{Codec Motion Fields}
\label{supp:implementation}

\subsection{Codec field construction}

We use signed codec measurements with native presentation timestamps, reference timestamps, source dimensions, and support masks.
The primary H.264 path reads native block-point records with quarter-pixel displacement.
A validated MPEG4 adapter preserves its corresponding signed displacement and temporal-reference convention.
Inter-prediction vectors exist on supported coded blocks, while P/B pictures can also contain intra-coded blocks.
Support masks distinguish unavailable motion from observed zero motion.
The potential uses signed motion measurements.

For a current timestamp $p\in\mathbb{Z}$, reference timestamp $r\in\mathbb{Z}$, native time base $\tau\in\R_{>0}$ seconds per tick, dimensions $W,H\in\N$, and native signed displacement $\mathbf{d}\in\R^2$, the H.264 image-normalized velocity is
\begin{equation}
 \mathbf{v}=\frac{\mathbf{d}}{4(r-p)\tau}\oslash(W,H),\qquad r\ne p,
\end{equation}
where $\oslash$ denotes component-wise division.
This sign convention follows the native current/reference relation.
Duplicate velocities at a current/reference/spatial point are averaged.
For each current/reference pair, we subtract the component-wise median over its observed spatial points and average the remaining reference-specific values at each point.
This deterministic compensation removes the shared translation of each frame pair.

We partition the native image into a grid of $C=4\times4$ cells.
For the scalar values $v_1,\ldots,v_d\in\R$ observed within one cell and component, the reducer is
\begin{equation}
 \rho(v_1,\ldots,v_d)=\frac{\sum_{h=1}^d v_h|v_h|}{\sum_{h=1}^d|v_h|}.
\end{equation}
When observed values are all zero, the result is zero.
When the cell has no supported observations, the result is missing.
Horizontal and vertical masks agree, which gives $2C$ channels in cell-wise paired order.
The full native timestamp sequence is retained for temporal widths and missingness.

\subsection{Question intervals and causal prefixes}

The public question defines the time domain before motion progress is normalized.
An instantaneous domain yields the neutral potential.
For SVCBench~\cite{svcbench}, we restrict both the current observation and every reference used to construct it to the query cutoff.
A field column is masked if its maximum used reference timestamp exceeds the cutoff.
Future reference information is therefore excluded even when the current coded frame precedes the query.
This reference-level filter accounts for prediction structures in which a displayed frame can reference a later timestamp.

For a prefix query at time $q\in\R_{\geq0}$, base RGB targets occur at $qi/15$ for $i\in\{0,\ldots,15\}$.
Each target selects the latest actual native frame at or before that timestamp.
All prefix images are causal, and the final public query sequence retains the official record order rather than being sorted by timestamp.
Codec fields are prepared offline and filtered for each query.
Inference runs per query, with no persistent incremental streaming state.

\section{Angular Count Evidence}

\subsection{Active axes and exact angular integration}

For one finite scalar component $\mathbf{y}\in\R^d$, let $\mathbf{X}\in\R^{d\times2}$ contain a constant column and normalized original source time.
We compute $r_0=\operatorname{rank}(\mathbf{X})\in\N_0$ and least-squares residual $\mathbf{u}=\mathbf{y}-\mathbf{X}\mathbf{X}^{\dagger}\mathbf{y}\in\R^d$.
The component is active when
\begin{equation}
 d-r_0>0,\quad
 \|\mathbf{u}\|_2^2>\epsilon\max(d,r_0)\|\mathbf{y}\|_2^2,
\end{equation}
where $\epsilon$ is FP64 machine precision and $\dagger$ denotes the least-squares pseudoinverse.
This baseline defines the active set and cell weights only.
The final candidate score uses angular phase evidence rather than the baseline regression fit.

For $d$ observed phases, the rotating-semicircle count changes only at their angles and their angles shifted by $\pi$, all taken modulo $2\pi$.
Sort the unique breakpoints together with 0 and $2\pi$.
The integral is the sum of each interval width divided by $2\pi$ times its constant beta-function term.
We evaluate the beta terms through log-gamma functions and the weighted sum through log-sum-exp.
No numerical quadrature or fitted angular concentration is required.
The two candidate rotation signs have fixed prior weights of one half.

The shared Euclidean motion clock is unchanged by a fixed rotation of the velocity coordinates.
This property concerns the clock: the full pipeline also includes component-wise reduction and active-axis decisions, both of which can change under rotation.

\subsection{Directional model behind the angular potential}
\label{supp:angular_model}

The semicircle statistic follows from a simple density on angles.
For $\mu\in[0,2\pi)$, let $H_\mu=(\mu,\mu+\pi]$ modulo $2\pi$, and let $\eta\in(0,1)$ be the probability mass assigned to that half.
The conditional density is
\begin{equation}
 f(\phi\mid\mu,\eta)=
 \begin{cases}
 \eta/\pi,&\phi\in H_\mu,\\
 (1-\eta)/\pi,&\phi\notin H_\mu.
 \end{cases}
 \label{eq:angular_density}
\end{equation}
For a fixed phase sequence $\boldsymbol{\phi}$ with $d$ observations, its product density is
$\pi^{-d}\eta^{h_{\boldsymbol{\phi}}(\mu)}(1-\eta)^{d-h_{\boldsymbol{\phi}}(\mu)}$.
Integrating $\eta$ against the symmetric $\operatorname{Beta}(a_0,a_0)$ prior yields
\begin{equation}
 \int_0^1
 \frac{\eta^{h+a_0-1}(1-\eta)^{d-h+a_0-1}}
      {\mathrm{B}(a_0,a_0)}\,\mathrm{d}\eta
 =\frac{\mathrm{B}(h+a_0,d-h+a_0)}
        {\mathrm{B}(a_0,a_0)}.
\end{equation}
Uniformly integrating $\mu$ gives $\pi^{-d}I(\boldsymbol{\phi})$.
The density is evaluated for the observed assignment of angles to the two halves, so no binomial coefficient enters.
For a fixed cell, all count hypotheses and the unwound reference use the same observations, so $\pi^{-d}$ cancels in their ratio.
Marginalizing the two rotation signs with equal weights gives the cell score in the main paper.

The model yields a directional compatibility statistic for each count.
\method uses it to condition the dVLM's count alternatives, so the visual-semantic interpretation of the model and the motion evidence act together on the answer.
Averaging the cell log ratios pools this evidence on a common scale across the frame.

For ideal straight-line reciprocation, write the fractional progress of a cycle as $r$.
The direction is $\theta_0$ for $0\leq r<1/2$ and $\theta_0+\pi$ for $1/2\leq r<1$.
Subtracting $2\pi r$ places both halves in the same semicircle $(\theta_0-\pi,\theta_0]$, apart from boundary points.
For ideal circular motion, subtracting the correct winding leaves a constant phase.
These examples show how one statistic serves both reversal and rotation.

\section{Structured Count Messages}

\subsection{Why normalize the count message?}

For the selected token support $\mathcal{S}$, the single-token correction satisfies
\begin{align}
 \sum_{a\in\mathcal{S}}e^{z'_a}
 &=e^{-\gamma}\sum_{a\in\mathcal{S}}e^{z_a+\ell(n(a))}\\
 &=\sum_{a\in\mathcal{S}}e^{z_a}.
\end{align}
All outside logits are unchanged, so the full softmax normalizer and total selected probability mass are preserved in exact arithmetic.
Constant potentials produce the original path.
The identity holds for the raw branch probabilities before guidance, and the BF16 implementation preserves it up to rounding.

\subsection{A single variational form for count messages}
\label{supp:variational}

Fix one denoising step and one condition branch.
Let $J$ be a block of one or two unresolved positions, and let
$P_J(\mathbf{a})=\prod_{j\in J}p_j(a_j)$ over $\mathcal{A}^{|J|}$.
Define $L_J(\mathbf{a})=\ell(n_J(\mathbf{a}))$ on the legal positive-count support $\mathcal{S}_J$ and $L_J=0$ elsewhere.
Consider normalized distributions in
\begin{equation}
 \mathcal{Q}_J=\left\{R\geq0:
 \sum_{\mathbf{a}}R(\mathbf{a})=1,\ 
 R(\mathbf{a})=P_J(\mathbf{a})\ 
 \text{for }\mathbf{a}\notin\mathcal{S}_J\right\}.
\end{equation}
The local update solves
\begin{equation}
 Q_J=\underset{R\in\mathcal{Q}_J}{\argmax}
 \left\{
 \sum_{\mathbf{a}}R(\mathbf{a})L_J(\mathbf{a})
 -\operatorname{KL}(R\|P_J)
 \right\}.
 \label{eq:variational}
\end{equation}
This form characterizes the inference update in closed form and involves no training.

The fixed complement forces
$\sum_{\mathcal{S}_J}R=M_J=\sum_{\mathcal{S}_J}P_J$.
For positive $P_J$ and nonempty support, stationarity with multiplier $\xi$ gives
\begin{equation}
 L_J(\mathbf{a})-\log\frac{R(\mathbf{a})}{P_J(\mathbf{a})}
 -1-\xi=0,\qquad \mathbf{a}\in\mathcal{S}_J.
\end{equation}
Thus $R(\mathbf{a})=P_J(\mathbf{a})e^{L_J(\mathbf{a})-1-\xi}$ on the support.
The mass constraint determines $e^{-1-\xi}=M_J/Z_J$, which recovers the main paper's update.
The objective is strictly concave on this finite positive support, so the solution is unique.
Empty supports take the identity path.
Finite raw logits give positive probabilities, and the implementation evaluates the update in log space.

Two useful properties follow directly.
First, a constant score on the support produces $Q_J=P_J$.
Second, adding any constant to all supported scores leaves the update unchanged.
Consequently, the unwound angular reference fixes the scale of the reported potential and cancels in the relative reweighting of counts.

The host expects factorized token distributions.
Writing $q_j$ for the marginals of $Q_J$, their product is the forward-KL projection
\begin{equation}
 \prod_{j\in J}q_j
 =\underset{\{r_j\}}{\argmin}\ 
 \operatorname{KL}\!\left(Q_J\,\middle\|\,\prod_{j\in J}r_j\right).
 \label{eq:factor_projection}
\end{equation}
Indeed, this objective differs from
$\sum_{j\in J}\operatorname{KL}(q_j\|r_j)$ only by a term independent of the $r_j$.
The log-ratio message realizes each projected marginal because
$e^{z_{j,a}}q_j(a)/p_j(a)=q_j(a)\sum_u e^{z_{j,u}}$.
For a pair, the projection keeps the coupled count evidence in each marginal and passes it to the host's per-position update.

\subsection{Integer messages across a digit and termination}
\label{supp:pair}

The main paper applies the count message jointly to the first unresolved numeric position $j$ before the terminator and an unresolved end-of-sequence (EOS) position $e$.
Let $p_j(a)$ and $p_e(b)$ be their raw full-vocabulary softmax probabilities in one condition branch.
The DFA defines the set $\mathcal{P}$ of token pairs whose completed drafts parse to positive integers $n(a,b)$.
We start from the local product $B(a,b)=p_j(a)p_e(b)$ and let $M=\sum_{(a,b)\in\mathcal{P}}B(a,b)$ denote its mass on positive integers.
The count-conditioned pair distribution is
\begin{equation}
 Q(a,b)=\begin{cases}
 \displaystyle\frac{M B(a,b)e^{\ell(n(a,b))}}
 {\sum_{(u,v)\in\mathcal{P}}B(u,v)e^{\ell(n(u,v))}},&(a,b)\in\mathcal{P},\\[3pt]
 B(a,b),&(a,b)\notin\mathcal{P}.
 \end{cases}
 \label{eq:pair}
\end{equation}
Like the single-token message, this tilt preserves the total mass $M$ of legal positive pairs and leaves the complement unchanged.
The host decoder scores each position separately, so we project $Q$ to its marginals $q_j(a)=\sum_bQ(a,b)$ and $q_e(b)=\sum_aQ(a,b)$ and update
\begin{equation}
 z'_{j,a}=z_{j,a}+\log\frac{q_j(a)}{p_j(a)},\qquad
 z'_{e,b}=z_{e,b}+\log\frac{q_e(b)}{p_e(b)}.
 \label{eq:marginal}
\end{equation}
The two exact marginals each normalize over the full vocabulary, and the host consumes them through its original token-update rule.
All other unresolved integer positions keep the single-token message, which also applies when the pair is unavailable or carries constant evidence.

The pair computation runs in FP64 on the logits' device and casts the updated logits back to the model's BF16 format.
We use excluded log-sum-exp computations for complement mass, avoiding unstable subtraction from one.
Only support token IDs, count mappings, and small pair tables enter Python-side planning, and full-vocabulary probabilities remain on the device.

\subsection{Tokenizer, terminator, and legal integer edits}

The integer draft contains all output slots, including committed content and unresolved raw-argmax proposals.
The DFA uses the actual parser grammar and token decoding of the host tokenizer, including the host's actual single-token terminator when one exists.
We enumerate legal one-token edits at each mutable row without deduplicating token aliases by integer value.
The two-variable domain uses the complete numeric and termination token domains derived from these neighbors and checks the full edited output with the DFA.
It covers every integer reachable from the current draft, independent of any label and without a fixed candidate range.

The pair update is applied to the first mutable atomic numeric position before the actual terminator and to the mutable EOS position.
Unavailable, singleton, and constant-evidence cases follow their applicable original or single-token path.
Hard-committed integer positions are never reopened.
For multiple choice, the complete public option-to-integer mapping and the tokenizer encoding of every option letter are verified before fusion.
Only the first answer row and valid positive numerical option letters receive the single-token update, and unsupported mappings keep the original decoding path.

\section{Denoising Integration}

\subsection{How count conditioning interacts with temporal contrast}
\label{supp:contrast}

The two visual branches use a common draft and edit support but different raw token probabilities.
For a unary support, let their normalization corrections be $\gamma_+$ and $\gamma_-$.
Before the plausibility filter, the guided logit of a supported token is
\begin{align}
 &(1+\alpha)z'_{+,a}-\alpha z'_{-,a}\nonumber\\
 &\quad=(1+\alpha)z_{+,a}-\alpha z_{-,a}
       +\ell(n(a))-\kappa,\\
 &\kappa=(1+\alpha)\gamma_+-\alpha\gamma_- .
\end{align}
The correction $\kappa$ is constant across the supported tokens.
For any two supported alternatives $a,u$, their pre-filter logit difference is therefore
\begin{align}
 \widehat z_a-\widehat z_u
 &= (1+\alpha)(z_{+,a}-z_{+,u})
    -\alpha(z_{-,a}-z_{-,u})\nonumber\\
 &\quad+\ell(n(a))-\ell(n(u)).
 \label{eq:guidance_difference}
\end{align}
Thus applying the same unary count evidence to both branches retains exactly one copy of its relative score after contrast.
The plausibility filter is computed from the conditioned positive logits, so the count evidence also shapes which tokens remain eligible.
For a pair, each branch derives its marginal messages from its own joint base distribution.

\subsection{Visual support and the shared denoising state}
\label{supp:full_loop}

Dream-VL's 16 base frames produce eight tubelets, each with $8\times8$ projected spatial features.
For supplementary observations, the fixed input plan uses available native times outside the base sample and assigns a total budget of at most 512 patches.
It distributes the budget across these times and ranks patches by the overlap-weighted codec motion magnitude in their spatial footprints, with a deterministic spatial-spread tie break.
Selection occurs after each supplementary frame has passed through the complete frozen vision encoder and projector, so it sets the visual-token budget of the language model while the encoder sees full frames.
The merged input retains spatial coordinates and chronological temporal ordering.

For the temporal gate, supported codec points are mapped to the base feature grid.
Let $u_{\tau r}$ be their mean camera-compensated speed pooled over the two frames of tubelet $\tau$ at location $r$.
The gate is $g_{\tau r}=u_{\tau r}/\max_{r'}u_{\tau r'}$ when the maximum is positive.
An unsupported or all-zero entry has gate zero.
Positions without codec support therefore keep their positive features in both conditions.
The gate is fixed before inference, as are the supplementary locations and their nearest base-tubelet anchors.
The negative condition thus attenuates temporal variation directly in feature space.

The complete inference order is:
\begin{enumerate}
 \item Prepare the query's motion coordinates, active set, visual memory, and two visual conditions.
 Initialize all $L$ output slots to MASK.
 \item Run both condition branches with the same committed tokens and unresolved-slot embeddings.
 Keep their visual-prefix and key/value states separate.
 \item Form the common draft from committed tokens and the positive branch's raw argmax.
 Build its unary and, when available, pair domains; obtain any uncached count scores.
 \item Apply the same domain and potential independently to both raw branches.
 Combine their updated logits by contrast and then the positive-branch plausibility filter.
 \item Construct the top-three soft proposal from these guided logits.
 Apply the native sampling and commitment schedule, including its existing termination adjustment.
 Hard commitments are permanent.
 \item Feed the soft proposal only to slots that remain unresolved in the next iteration, sharing it across both branches.
 Repeat until the $K$-step schedule has completed.
\end{enumerate}
Soft proposals are computed before the native sampler's termination adjustment.
For unit vectors $\mathbf{u},\mathbf{v}$ with angle $\omega=\arccos(\mathbf{u}^{\mathsf T}\mathbf{v})\in(0,\pi)$,
\begin{equation}
 \operatorname{slerp}(\mathbf{u},\mathbf{v};\lambda)
 =\frac{\sin((1-\lambda)\omega)}{\sin\omega}\mathbf{u}
  +\frac{\sin(\lambda\omega)}{\sin\omega}\mathbf{v}.
\end{equation}
The implementation renormalizes this direction and restores the MASK embedding norm.
A zero proposal mean retains MASK.
Nearly parallel vectors use normalized linear interpolation, and the antipodal case uses a deterministic orthogonal great-circle direction.
A zero MASK embedding stays zero.

With supplementary features, equal positive and negative feature arrays trigger the identity guidance path.
Without supplementary features, the inherited zero-gate rule triggers it.
The implementation represents the guidance mask by the smallest finite BF16 value, numerically suppressing blocked alternatives.
Potential evaluation and token-domain planning require no model forward passes.
The two visual branches and supplementary image encoding are costs of the full Dream system and are retained in its matched ablation.

\end{document}